\documentclass[11pt]{article}
\usepackage{colacl}
\title{Towards an implementable dependency grammar\thanks{in CoLing-ACL'98 workshop 
	Processing of Dependency-based Grammars, Kahane and Polgu{\`e}re (eds), p. 1-10, Montreal, Canada, 1998}}
\author{Timo J{\"a}rvinen \and Pasi Tapanainen \\
  Research Unit for Multilingual Language Technology \\
  P.O. Box 4, FIN-00014 University of Helsinki, Finland}

\newenvironment{acknowledgments}{\small\par{\bf\noindent Acknowledgments}\par\vspace{2pt}\raggedright}{\par}

\begin{document}
\maketitle
\begin{abstract}
  Syntactic models should be descriptively adequate and parsable.  A
  syntactic description is autonomous in the sense that it has certain
  explicit formal properties.  Such a description relates to the
  semantic interpretation of the sentences, and to the surface text.
  As the formalism is implemented in a broad-coverage syntactic
  parser, we concentrate on issues that must be resolved by any
  practical system that uses such models.  The correspondence between
  the structure and linear order is discussed.
\end{abstract}

\section{Introduction}
\label{sec:intro}

The aim of this paper is to define a dependency grammar
framework which is both linguistically motivated and computationally
parsable.

A linguistically adequate grammar is the primary target because if we
fail to define a descriptive grammar, its application is less useful
for any linguistically motivated purposes. In fact, our
understanding of the potential benefits of the linguistic means can
increase only if our practical solutions stand on an adequate
descriptive basis.

Traditionally, grammatical models have been constructed by
linguists without any consideration for computational application, and
later, by computationally oriented scientists who have first taken a
parsable mathematical model and then forced the linguistic description
into the model which has usually been too weak to describe what a linguist
would desire.

Our approach is somewhere between these two extremes.  While we define
the grammar strictly in linguistic terms, we simultaneously test it in
the parsing framework.  What is exceptional here is that the parsing
framework is not restricted by an arbitrary mathematical model such as
a context-free phrase structure grammar.  This leads us to a situation
where the parsing problem is extremely hard in the general,
theoretical case, but fortunately parsable in practise.  Our result
shows that, while in general we have an NP-hard parsing problem, there
is a specific solution for the given grammar that can be run quickly.
Currently, the speed of the parsing system\footnote{Demo: \em
http://www.conexor.fi/analysers.html} is several hundred words per
second.

In short, the grammar should be empirically motivated.  We have all the
reason to believe that if a linguistic analysis rests on a solid
descriptive basis, analysis tools based on the theory would be more
useful for practical purposes.  We are studying the possibilities of
using computational implementation as a developing and testing
environment for a grammatical formalism.  We refer to the
computational implementation of a grammar as a {\em parsing grammar}.

\subsection{Adequacy}
\label{sec:adeq}

A primary requirement for a parsing grammar is that it is
descriptively adequate. Extreme distortion results if the mathematical
properties of the chosen model restrict the data.  However, this
concern is not often voiced in the discussion. For example,
\newcite[p.~92]{JDMc82} notes that such a basic assumption concerning
linguistic structures that ``{\em strings are more basic than trees and
that trees are available only as a side product of derivations that
operate in terms of strings}'' was attributable to {\em the historical
accident that early transformational grammarians knew some automata
theory but no graph theory}.''

One reason for computationally oriented syntacticians to favour
restricted formalisms is that they are easier to implement.  Those who
began to use dependency models in the 1960's largely ignored
descriptive adequacy in order to develop models which were
mathematically simple and, as a consequence, for which effective
parsing algorithms could be presented.  These inadequacies had to be
remedied from the beginning, which resulted in ad hoc theories or
engineering solutions\footnote{See discussion of an earlier
  engineering art in applying a dependency grammar in
  \newcite{KiKe94}.} without any motivation in the theory.

There have been some serious efforts to resolve these problems.
Hudson~\shortcite{RH89b}, for example, has attempted to construct a
parser that would reflecs the claims of the theory (Word Grammar) as
closely as possible.  However, it seems that even linguistically
ambitious dependency theories, such as Hudson's Word Grammar, contain
some assumptions which are attributable to certain mathematical
properties of an established formalism rather than imposed by the
linguistic data\footnote{For instance, the notion of adjacency was 
redefined in WG, but was still unsuitable for ``free'' word order languages.}.  These
kinds of unwarranted assumptions tend to focus the discussion on
phenomena which are rather marginal, if a complete description of a
language is concerned.  No wonder that comprehensive descriptions,
such as \newcite{Qu85}, have usually been non-formal.

\subsection{The European structuralist tradition}
\label{sec:empirical}

We argue for a syntactic description that is based on dependency
rather than constituency, and we fully agree with \newcite[p.~1]{EH93}
that {\em ``making use of the presystemic insights of classical
  European linguistics, it is then possible that constituents may be
  dispensed with as basic elements of (the characterization of) the sentence
  structure.''}  However, we disagree with the notion of
``presystemic'' if it is used to imply that earlier work is obsolete.
 From a descriptive point of view, it is crucial to look at the data
that was covered by earlier non-formal grammarians.

As far as syntactic theory is concerned, there is no need to reinvent
the wheel.  Our description has its basis in the so-called ``classical
model'' based on the work of the French linguist Lucien Tesni{\`e}re.  His
structural model should be capable of describing any occurring natural
language.  His main work, \shortcite{LT59} addresses a large amount of
material from typologically different languages.  It is indicative of
Tesni{\`e}re's empirical orientation that there are examples from some 60
languages, though his method was not empirical in the sense that he
would have used external data inductively.  As \newcite{HJH96} points
out, Tesni{\`e}re used data merely as an expository device. However, in
order to achieve formal rigour he developed a model of syntactic
description, which obviously stems from the non-formal tradition
developed since antiquity but without compromising the descriptive
needs.  We give a brief historical overview of the formal properties
inherent in Tesni{\`e}re's theory in Section~\ref{sec:formal} before we
proceed to the implementational issues in Section~\ref{sec:system}.

\subsection{The surface syntactic approach}
\label{sec:surf}

We aim at a theoretical framework where we have a dependency theory
that is both descriptively adequate and formally explicit.  The latter
is required by the broad-coverage parsing grammar for English that we
have implemented.  We maintain the parallelism between the syntactic
structure and the semantic structure in our design of the syntactic
description: when a choice between alternative syntactic constructions
in a specific context should be made, the semantically motivated
alternative is selected\footnote{In such sentence as {\em ``I asked
    John to go home''}, the noun before the infinitive clause is
  analysed as the (semantic) subject of the infinitive rather than as
  a complement of the governing verb.}.

Although semantics determines what kind of structure a certain
sentence should have, from the practical point of view, we have a
completely different problem: how to resolve the syntactic structure
in a given context.  Sometimes, the latter problem leads us back to
redefine the syntactic structure so that it can be detected in the
sentence\footnote{For instance, detecting the distinct roles of the
  {\em to}-infinitive clause in the functional roles of the purpose or
  reason is usually difficult (e.g.~\newcite[p.~564]{Qu85}: {\em ``Why
    did he do it?}; purpose: {\em ``To relieve his anger''} and
  reason: {\em ``Because he was angry''}).  In such sentence as {\em
    ``A man came to the party to have a good time''}, the
  interpretation of the infinitive clause depends on the interaction
  of the contextual and lexical semantics rather than a structural
  distinction.}.  Note, however, that this redefinition is now made on
a linguistic basis.  In order to achieve parsability, the surface
description should not contain elements which can not be selected by
using contextual information.  It is important that the redefinition
should not be made because an arbitrary mathematical model denies
e.g.~crossing dependencies between the syntactic elements.

\section{Constituency vs.~dependency}
\label{sec:constit}

A central idea in American structuralism was to develop rigorous
mechanical procedures, i.e.~``discovery procedures'', which were
assumed to decrease the grammarians' own, subjective assessment in the
induction of the grammars.  This practice was culminated in
\newcite[p.~5]{ZH60}, who claimed that {\em ``the main research of
  descriptive linguistics, and the only relation which will be
  accepted as relevant in the present survey, is the distribution or
  arrangement within the flow of speech of some parts or features
  relative to others.''}

The crucial descriptive problem for a distributional grammar
(i.e.~phrase-structure grammar) is the existence of non-contiguous
elements.  The descriptive praxis of some earlier IC theoricians
allows discontiguous constituents.  For example, already
\newcite{RW47} discussed the problem at length and defined a
restriction for discontiguous constituents\footnote{\newcite{RW47}:
  {\em ``A discontinuous sequence is a constituent if in some
    environment the corresponding continuous sequence occurs as a
    constituent in a construction semantically harmonious with the
    constructions in which the given discontinuous sequence occurs.''}
  Further, Wells notes that ``The phrase {\em semantically harmonious}
  is left undefined, and will merely be elucidated by examples.''}.
Wells' restriction implies that a discontiguous sequence can be a
constituent only if it appears as a contiguous sequence in another
context.  This means that Wells' characterisation of a constituent
defines an element which is broadly equivalent to the notion of bunch
in Tesni{\`e}re's~\shortcite{LT59} theory.  Consequently, these two types
of grammars are capable of describing the equivalent syntactic
phenomena and share the assumption that a syntactic structure is
compatible with its semantic interpretation.  However, the extended
constituent grammar thus no longer provides a rigorous distributional
basis for a description, and its formal properties are unknown.

We can conclude our argument by stating that the reason to reject
constitutional grammars is that the formal properties for
descriptively adequate constitutional grammars are not known.  In the
remaining sections, we show that a descriptively adequate dependency
model can be constructed so that it is formally explicit and
parsable.

\section{Parallelism between the syntactic and semantic structures}
\label{sec:synsem}

Obviously, distributional descriptions that do not contribute to
their semantic analysis can be given to linguistic strings.
Nevertheless, the minimal descriptive requirement should be that a
syntactic description is compatible with the semantic structure.  The
question which arises is that if the correspondence between syntactic
and semantic structures exists, why should these linguistic levels be
separated.  For example, \newcite[p.~278]{PSg92} has questioned the
necessity of the syntactic level altogether.  His main argument for
dispensing with the whole surface syntactic level is that there are no
strictly synonymous syntactic constructions, and he therefore suggests
that the surface word order belongs more properly to the level of
morphemics.  This issue is rather complicated.  We agree that surface
word order does not belong to syntactic structure, but for different
reasons.

In contradistinction to Sgall's claim, \newcite[p.~33]{IM87} has
provided some evidence where the morphological marker appears either
in the head or the dependent element in different languages, as in the
Russian {\em ``kniga professor+a''} (professor's book) and its
Hungarian equivalent {\em ``professzor k{\"o}nyv+e''}.  Consequently,
\newcite[p.~108]{IM87} distinguishes the morphological dependency as a
distinct type of dependency.  Thus morphology does not determine the
syntactic dependency, as \newcite[Ch.~15]{LT59} also argues.

For \newcite[Ch.~20:17]{LT59} meaning (Fr.~{\em sens}) and
structure are, in principle, independent.  This is backed by the
intuition that one recognises the existence of the linguistic
structures which are semantically absurd, as illustrated by the
structural similarity between the nonsensical sentence {\em ``Le
  silence vertebral indispose la voie licite''} and the meaningful
sentence {\em ``Le signal vert indique la voie libre''}.

The independence of syntactic and semantic levels is crucial for
understanding Tesni{\`e}re's thesis that the syntactic structure follows
 from the semantic structure, but not vice versa.  This means that
whenever there is a syntactic relation, there is a semantic relation
(e.g.~complementation or determination) going in the opposite
direction.  In this view, the syntactic head requires semantic
complementation from its dependents.  Only because the syntactic and
semantic structures belong to different levels is there no
interdependency or mutual dependency, though the issue is sometimes
raised in the literature.

There is no full correspondence between the syntactic and semantic
structures because some semantic relations are not marked in the
functional structure.  In \newcite[p.~85]{LT59}, for example, there
are anaphoric relations, semantic relations without correspondent
syntactic relations.

\section{Surface representation and syntactic structure}

\subsection{The nucleus as a syntactic primitive}

The dependency syntactic models are inherently more ``word oriented''
than constituent-structure models, which use abstract phrase
categories.  The notion of word, understood as an orthographic unit in
languages similar to English, is not the correct choice as a syntactic
primitive. However, many dependency theories assume that the
orthographic words directly correspond\footnote{See
  \newcite[p.~491]{JK75} and \newcite{RH91}.} to syntactic primitives
(nodes in the trees).  Although the correspondence could be very close
in languages like English, there are languages where the word-like
units are much longer (i.e.~incorporating languages).

Tesni{\`e}re observed that because the syntactic connexion implies a
parallel semantic connexion, each node has to contain a syntactic and
a semantic centre.  The node element, or {\em nucleus}, is the
genuine syntactic primitive.  There is no one-to-one correspondence
between nuclei and orthographic words, but the nucleus consists of one
or more, possibly discontiguous, words or parts of words.  The
segmentation belongs to the linearisation, which obeys
language-specific rules.  Tesni{\`e}re \shortcite[Ch~23:17]{LT59} argued
that the notion {\em word}, a linear unit in a speech-chain, does not
belong to syntactic description at all.  A {\em word} is nothing
but a {\em segment} in the speech chain \shortcite[Ch~10:3]{LT59}.

The basic element in syntactic description is the nucleus.  It
corresponds to a node in a dependency tree.  When the sentence is
represented as a dependency tree, the main node contains the whole
verb chain.

There are at least two reasons why the concept of the nucleus is
needed. In the first place, there are no cross-linguistically valid
criteria to determine the head in, say, a prepositional
phrase.  One may decide, arbitrarily, that either the preposition or
the noun is the head of the construction.  Second, because the nucleus
is also the basic semantic unit, it is the minimal unit in a
lexicographical description.

\subsection{Linearisation}

Tesni{\`e}re makes a distinction between the {\em linear order}, which is
a one-dimensional property of the physical manifestations of the
language, and the {\em structural order}, which is two-dimensional.
According to his conception, constructing the structural description
is converting the linear order into the structural order.  Restricting
himself to syntactic description, Tesni{\`e}re does not formalise this
conversion though he gives two main principles: (1) usually dependents
either immediately follow or precede their heads (projectivity) and
when they do not, (2) additional devices such as
morphological agreement can indicate the connexion.

Although Tesni{\`e}re's distinction between the linear and structural
order corresponds to some extent with the distinction between the
linear precedence (LP) and the immediate dominance, there is a crucial
difference in emphasis with respect to those modern syntactic
theories, such as GPSG, that have distinct ID and LP components.
Tesni{\`e}re excludes word order phenomena from his structural syntax and
therefore does not formalise the LP component at all.  Tesni{\`e}re's
solution is adequate, considering that in many languages word order is
considerably free.  This kind of ``free'' word order means that the
alternations in the word order do not necessarily change the meaning
of the sentence, and therefore the structural description implies
several linear sequences of the words.  This does not mean that there
are no restrictions in the linear word order but these restrictions do
not emerge in the structural analysis.

In fact, Tesni{\`e}re assumes that a restriction that is later formalised
as an {\em adjacency principle} characterizes the neutral word order
when he says that there are no syntactic reasons for violating
adjacency in any language, but the principle can be violated, as he
says, for {\em stylistic reasons} or to save the metric structure in
poetics.  If we replace the stylistic reasons with the more broader
notion which comprises the discourse functions, his analysis seems
quite consistent with our view.  Rather than seeing that there are
syntactic restrictions concerning word order, one should think that
some languages due to their rich morphology have more freedom in using
word order to express different discourse functions.  Thus,
linearisation rules are not formal restrictions, but language-specific
and functional.

There is no need for constituents.  Tesni{\`e}re's theory has two
mechanisms to refer to linearisation.  First, there are static
functional categories with dynamic potential to change the initial
category.  Thus, it is plausible to separately define the
combinatorial and linearisation properties of each category.  Second,
the categories are hierarchical so that, for instance, a verb in a
sentence governs a noun, an adverb or an adjective.  The lexical
properties, inherent to each lexical element, determine what the
governing elements are and what elements are governed.

There are no simple rules or principles for linearisation.
Consider, for example, the treatment of adjectives in English.  The
basic rule is that attributive adjectives precede their heads.
However, there are notable exceptions, including
the postmodified adjectives\footnote{Example: {\em ``It is a
    phenomenon \underline{consistent} with $\ldots$''}}, which follow their
heads, and some lexical exceptions\footnote{Example: {\em ``president
    \underline{elect}''}}, which usually or always are postmodifying.

\section{Historical formulations}
\label{sec:formal}

In this section, the early formalisations of the dependency grammar
and their relation to Tesni{\`e}re's theory are discussed.  The dependency
notion was a target of extensive formal studies already in the first
half of the 1960's\footnote{A considerable number of the earlier
  studies were listed by \newcite[p.~263]{SM67}, who also claimed that
  {\em ``Tesni{\`e}re was one of the first who used (dependency) graphs in
    syntax.  His ideas were repeated, developed and precised by
    Y.~Lecerf \& P.~Ihm (1960), L.~Hirschberg and I.~Lynch,
    particularly by studying syntactic projectivity and linguistic
    subtrees.''}}.

\subsection{Gaifman's formulation}
\label{sec:gaif}

The classical studies of the formal properties of dependency grammar
are \newcite{HG65} and \newcite{Hays64}\footnote{Tesni{\`e}re is not
  mentioned in these papers.  Gaifman's paper describes the results
  ``$\ldots$ obtained while the author was a consultant for the RAND
  Corporation in the summer of 1960.''  Whereas phrase-structure
  systems were defined by referring to Chomsky's Syntactic Structures,
  the corresponding definition for the dependency systems reads as follows:
  ``By dependency system we mean a system, containing a finite number
  of rules, by which dependency analysis for certain language is done,
  as described in certain RAND publications (Hays, February 1960; Hays
  and Ziehe, April 1960).''  Speaking of the dependency theory,
  \newcite{Hays60} refers to the Soviet work on machine
  translation using the dependency theory of Kulagina et~al.  In
  \newcite{Hays64}, the only linguistic reference is to the 1961 edition
  of Hjelmslev's Prolegomena: {\em ``Some of Hjelmslev's empirical
    principles are closely related to the insight behind dependency
    theory, but empirical dependency in his sense cannot be identified
    with abstract dependency in the sense of the present paper, since
    he explicitly differentiates dependencies from other kinds of
    relations, whereas the present theory intends to be complete,
    i.e.~to account for all relations among units of utterances.''}},
which demonstrate that dependency grammar of the given type is weakly
equivalent to the class of context-free phrase-structure grammars.
The formalisation of dependency grammars is given in
\newcite[p.~305]{HG65}: For each category $X$, there will be a finite
number of rules of the type $X(Y_{1}, Y_{2} \cdots Y_{l} * Y_{l+1}
\cdots Y_{n})$, which means that $Y_{1} \cdots Y_{n}$ can depend on
$X$ in this given order, where $X$ is to occupy the position of $*$.

Hays, referring to Gaifman's formulation above, too strongly claims
that {\em ``[d]ependency theory is weakly omnipotent to IC theory. The
  proof is due to Gaifman, and is too lengthy to present here.  The
  consequence of Gaifman's theorem is that the class of sets of
  utterances [...]  is Chomsky's class of context-free languages.''}
This claim was later taken as granted to apply to any dependency
grammar, and the first, often cited, attestation of this apparently
false claim appeared in \newcite{Lg46}.  She presented four axioms of
the theory and claimed they were advocated by Tesni{\`e}re and formalised
by Hays and Gaifman.

Thus, the over-all result of the Gaifman-Hays proof was that there is
a weak equivalence of dependency theory and context-free
phrase-structure grammars.  This weak equivalence means only that both
grammars characterize the same sets of strings.  Unfortunately, this
formulation had little to do with Tesni{\`e}re's dependency theory, but as
this result met the requirements of a characterisation theory,
interest in the formal properties of dependency grammar diminished
considerably.

\subsection{Linguistic hypotheses}
\label{sec:hypos}

{\em Tesni{\`e}re's Hypothesis}, as \newcite{SM67} calls it, assumes that
each element has exactly one head.  Marcus also formulates a stronger
hypothesis, the {\em Projectivity hypothesis}, which connects the
linear order of the elements of a sentence to the structural order of
the sentence.  The hypothesis is applied in the following formulation:
let $x = a_{1}a_{2} \ldots a_{i} \ldots a_{n}$ be a sentence, where
$a_{i}$ and $a_{j}$ are terms in the sentence.  If the term $a_{i}$ is
subordinate to the term $a_{j}$, and there is an index $k$ which holds
$min(i,j) < k < max(i,j)$, then the term $a_{k}$ is subordinate to the
term $a_{j}$.

This is the formal definition of projectivity, also known as {\em
  adjacency} or {\em planarity}.  The intuitive content of adjacency
is that modifiers are placed adjacent to their heads.  The intuitive
content behind this comes from Behaghel's First Law\footnote{\em ``The
  most important law is that what belongs together mentally
  (semantically) is placed close together syntactically.''}
\cite[p.~143]{AS88}.

The adjacency principle is applicable only if the linear order of
strings is concerned.  However, the target of Tesni{\`e}re's syntax is
structural description and, in fact, Tesni{\`e}re discusses linear order,
a property attributable to strings, only to exclude linearisation from
his conception of syntax.  This means that a formalisation which
characterises sets of strings can not even be a partial formalisation
of Tesni{\`e}re's theory because his syntax is not concerned with strings,
but structures.  Recently, \newcite{PNNB97} have studied some formal
properties of dependency grammar, observing that Gaifman's conception
is not compatible either with Tesni{\`e}re's original formulation or with
the ``current'' variants of DG.

\begin{figure}[t!]
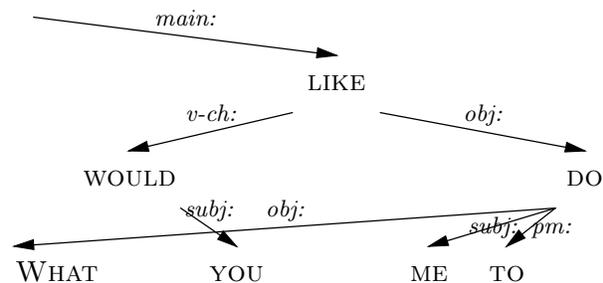

  \begin{center}
    \leavevmode
    \expandafter\ifx\csname graph\endcsname\relax \csname newbox\endcsname\graph\fi
    \expandafter\ifx\csname graphtemp\endcsname\relax \csname newdimen\endcsname\graphtemp\fi
    \setbox\graph=\vtop{\vskip 0pt\hbox{%
        \graphtemp=.5ex\advance\graphtemp by 0.150in
        \rlap{\kern 0.231in\lower\graphtemp\hbox to 0pt{\hss {\sc }\hss}}%
        \graphtemp=.5ex\advance\graphtemp by 1.650in
        \rlap{\kern 0.221in\lower\graphtemp\hbox to 0pt{\hss {\sc ~~~~~~~~What}\hss}}%
        \graphtemp=.5ex\advance\graphtemp by 1.150in
        \rlap{\kern 0.831in\lower\graphtemp\hbox to 0pt{\hss {\sc would}\hss}}%
        \graphtemp=.5ex\advance\graphtemp by 1.650in
        \rlap{\kern 1.392in\lower\graphtemp\hbox to 0pt{\hss {\sc you}\hss}}%
        \special{pn 8}%
        \special{pa 1100 1300}%
        \special{pa 1392 1500}%
        \special{fp}%
        \special{sh 1.000}%
        \special{pa 1324 1423}%
        \special{pa 1392 1500}%
        \special{pa 1296 1464}%
        \special{pa 1324 1423}%
        \special{fp}%
        \graphtemp=\baselineskip\multiply\graphtemp by -1\divide\graphtemp by 2
        \advance\graphtemp by .5ex\advance\graphtemp by 1.400in
        \rlap{\kern 1.246in\lower\graphtemp\hbox to 0pt{\hss {\footnotesize\em subj:}\hss}}%
        \graphtemp=.5ex\advance\graphtemp by 0.650in
        \rlap{\kern 1.915in\lower\graphtemp\hbox to 0pt{\hss {\sc like}\hss}}%
        \special{pa 331 300}%
        \special{pa 1915 500}%
        \special{fp}%
        \special{sh 1.000}%
        \special{pa 1819 463}%
        \special{pa 1915 500}%
        \special{pa 1813 512}%
        \special{pa 1819 463}%
        \special{fp}%
        \graphtemp=\baselineskip\multiply\graphtemp by -1\divide\graphtemp by 2
        \advance\graphtemp by .5ex\advance\graphtemp by 0.400in
        \rlap{\kern 1.123in\lower\graphtemp\hbox to 0pt{\hss {\footnotesize\em main:}\hss}}%
        \special{pa 1685 800}%
        \special{pa 831 1000}%
        \special{fp}%
        \special{sh 1.000}%
        \special{pa 934 1002}%
        \special{pa 831 1000}%
        \special{pa 922 953}%
        \special{pa 934 1002}%
        \special{fp}%
        \graphtemp=\baselineskip\multiply\graphtemp by -1\divide\graphtemp by 2
        \advance\graphtemp by .5ex\advance\graphtemp by 0.900in
        \rlap{\kern 1.258in\lower\graphtemp\hbox to 0pt{\hss {\footnotesize\em v-ch:}\hss}}%
        \graphtemp=.5ex\advance\graphtemp by 1.650in
        \rlap{\kern 2.400in\lower\graphtemp\hbox to 0pt{\hss {\sc me}\hss}}%
        \graphtemp=.5ex\advance\graphtemp by 1.650in
        \rlap{\kern 2.808in\lower\graphtemp\hbox to 0pt{\hss {\sc to}\hss}}%
        \graphtemp=.5ex\advance\graphtemp by 1.150in
        \rlap{\kern 3.215in\lower\graphtemp\hbox to 0pt{\hss {\sc do}\hss}}%
        \special{pa 2146 800}%
        \special{pa 3215 1000}%
        \special{fp}%
        \special{sh 1.000}%
        \special{pa 3122 957}%
        \special{pa 3215 1000}%
        \special{pa 3112 1006}%
        \special{pa 3122 957}%
        \special{fp}%
        \graphtemp=\baselineskip\multiply\graphtemp by -1\divide\graphtemp by 2
        \advance\graphtemp by .5ex\advance\graphtemp by 0.900in
        \rlap{\kern 2.681in\lower\graphtemp\hbox to 0pt{\hss {\footnotesize\em obj:}\hss}}%
        \special{pa 3062 1300}%
        \special{pa 231 1500}%
        \special{fp}%
        \special{sh 1.000}%
        \special{pa 332 1518}%
        \special{pa 231 1500}%
        \special{pa 329 1468}%
        \special{pa 332 1518}%
        \special{fp}%
        \graphtemp=\baselineskip\multiply\graphtemp by -1\divide\graphtemp by 2
        \advance\graphtemp by .5ex\advance\graphtemp by 1.400in
        \rlap{\kern 1.646in\lower\graphtemp\hbox to 0pt{\hss {\footnotesize\em obj:}\hss}}%
        \special{pa 3062 1300}%
        \special{pa 2400 1500}%
        \special{fp}%
        \special{sh 1.000}%
        \special{pa 2503 1495}%
        \special{pa 2400 1500}%
        \special{pa 2488 1447}%
        \special{pa 2503 1495}%
        \special{fp}%
        \graphtemp=.5ex\advance\graphtemp by 1.400in
        \rlap{\kern 2.731in\lower\graphtemp\hbox to 0pt{\hss {\footnotesize\em subj:}\hss}}%
        \special{pa 3062 1300}%
        \special{pa 2808 1500}%
        \special{fp}%
        \special{sh 1.000}%
        \special{pa 2902 1458}%
        \special{pa 2808 1500}%
        \special{pa 2871 1418}%
        \special{pa 2902 1458}%
        \special{fp}%
        \graphtemp=.5ex\advance\graphtemp by 1.400in
        \rlap{\kern 2.935in\lower\graphtemp\hbox to 0pt{{\footnotesize\em pm:}\hss}}%
        \graphtemp=.5ex\advance\graphtemp by 1.650in
        \hbox{\vrule depth1.800in width0pt height 0pt}%
        \kern 3.777in
        }%
      }%
    \centerline{\raise 1em\box\graph}
    \caption{Non-projective dependency tree}
    \label{fig:whatdo}
  \end{center}
\end{figure}

There are several equivalent formalisations for this intuition. In effect
they say that in a syntactic tree, where words are printed in
linear order, the arcs between the words must not cross.  For example,
in our work, as the arc between the node ``what'' and the node ``do''
in Figure~\ref{fig:whatdo} violates the principle, the construction is
non-projective.

\subsection{Formal properties of a Tesni{\`e}re-type DG}

Our current work argues for a dependency grammar that is conformant
with the original formulation in \newcite{LT59} and contains the
following axioms:
\begin{itemize}
\item The primitive element of syntactic description is a nucleus.
\item Syntactic structure consists of connexions between nuclei.
\item Connexion \cite[Ch.~1:11]{LT59} is a binary functional relation
  between a superior term (regent) and inferior term (dependent).
\item Each nucleus is a node in the syntactic tree and it has exactly
  one regent \cite[Ch.~3:1]{LT59}.
\item A regent, which has zero or more dependents, represents the whole
  subtree.
\item The uppermost regent is the central node of the sentence.
\end{itemize}

These axioms define a structure graph which is acyclic and
directed, i.e.~the result is a tree.  These strong empirical
claims restrict the theory.  For example, multiple dependencies
and all kinds of cyclic dependencies, including mutual dependency, are
excluded.  In addition, there can be no isolated nodes.

However, it is not required that the structure be projective, a
property usually required in many formalised dependency theories that
do not take into account the empirical fact that non-projective
constructions occur in natural languages.

\section{The Functional Dependency Grammar}
\label{sec:system}

Our parsing system, called the Functional Dependency Grammar (FDG),
contains the following parts:
\begin{itemize}
\item the lexicon,
\item the CG-2 morphological disambiguation \cite{AVo95,PTa96}, and
\item the Functional Dependency Grammar \cite{PTaTJa97,TJaPTa97}.
\end{itemize}

\subsection{On the formalism and output}

It has been necessary to develop an expressive formalism to represent
the linguistic rules that build up the dependency structure.  The
descriptive formalism developed by Tapanainen can be used to write
effective recognition grammars and has been used to write
a comprehensive parsing grammar of English.

When doing fully automatic parsing it is necessary to address
word-order phenomena.  Therefore, it is necessary that the grammar
formalism be capable of referring simultaneously both to syntactic
order and linear order.  Obviously, this feature is an extension of
Tesni{\`e}re's theory, which does not formalise linearisation.  Our
solution, to preserve the linear order while presenting the
structural order requires that functional information is no longer
coded to the canonical order of the dependents\footnote{Compare  this solution
with
  the Prague approach, which uses horizontal ordering as a formal
  device to express the topic-focus articulation at their
  tectogrammatical level.  The mapping from the tectogrammatical level
  to the linear order requires separate rules, called {\em shallow
    rules} \cite{VP87}.  Before such a description exists, one can not
  make predictions concerning the complexity of the grammar.}.

In the FDG output, the functional information is represented
explicitly using arcs with labels of syntactic functions.
Currently, some 30 syntactic functions are applied.

\begin{figure}[tb]
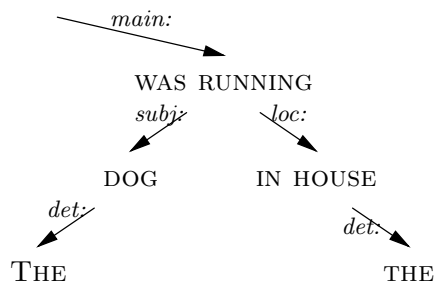

  \begin{center}
    \leavevmode
    \expandafter\ifx\csname graph\endcsname\relax \csname newbox\endcsname\graph\fi
    \expandafter\ifx\csname graphtemp\endcsname\relax \csname newdimen\endcsname\graphtemp\fi
    \setbox\graph=\vtop{\vskip 0pt\hbox{%
        \graphtemp=.5ex\advance\graphtemp by 1.650in
        \rlap{\kern 0.192in\lower\graphtemp\hbox to 0pt{\hss {\sc The}\hss}}%
        \graphtemp=.5ex\advance\graphtemp by 1.150in
        \rlap{\kern 0.677in\lower\graphtemp\hbox to 0pt{\hss {\sc dog}\hss}}%
        \special{pn 8}%
        \special{pa 485 1300}%
        \special{pa 192 1500}%
        \special{fp}%
        \special{sh 1.000}%
        \special{pa 289 1464}%
        \special{pa 192 1500}%
        \special{pa 261 1423}%
        \special{pa 289 1464}%
        \special{fp}%
        \graphtemp=\baselineskip\multiply\graphtemp by -1\divide\graphtemp by 2
        \advance\graphtemp by .5ex\advance\graphtemp by 1.400in
        \rlap{\kern 0.338in\lower\graphtemp\hbox to 0pt{\hss {\footnotesize\em det:}\hss}}%
        \graphtemp=.5ex\advance\graphtemp by 0.650in
        \rlap{\kern 1.162in\lower\graphtemp\hbox to 0pt{\hss {\sc was running}\hss}}%
        \special{pa 292 300}%
        \special{pa 1162 500}%
        \special{fp}%
        \special{sh 1.000}%
        \special{pa 1070 453}%
        \special{pa 1162 500}%
        \special{pa 1058 502}%
        \special{pa 1070 453}%
        \special{fp}%
        \graphtemp=\baselineskip\multiply\graphtemp by -1\divide\graphtemp by 2
        \advance\graphtemp by .5ex\advance\graphtemp by 0.400in
        \rlap{\kern 0.727in\lower\graphtemp\hbox to 0pt{\hss {\footnotesize\em main:}\hss}}%
        \special{pa 969 800}%
        \special{pa 677 1000}%
        \special{fp}%
        \special{sh 1.000}%
        \special{pa 774 964}%
        \special{pa 677 1000}%
        \special{pa 745 923}%
        \special{pa 774 964}%
        \special{fp}%
        \graphtemp=\baselineskip\multiply\graphtemp by -1\divide\graphtemp by 2
        \advance\graphtemp by .5ex\advance\graphtemp by 0.900in
        \rlap{\kern 0.823in\lower\graphtemp\hbox to 0pt{\hss {\footnotesize\em subj:}\hss}}%
        \graphtemp=.5ex\advance\graphtemp by 1.150in
        \rlap{\kern 1.646in\lower\graphtemp\hbox to 0pt{\hss {\sc in house}\hss}}%
        \special{pa 1354 800}%
        \special{pa 1646 1000}%
        \special{fp}%
        \special{sh 1.000}%
        \special{pa 1578 923}%
        \special{pa 1646 1000}%
        \special{pa 1550 964}%
        \special{pa 1578 923}%
        \special{fp}%
        \graphtemp=\baselineskip\multiply\graphtemp by -1\divide\graphtemp by 2
        \advance\graphtemp by .5ex\advance\graphtemp by 0.900in
        \rlap{\kern 1.500in\lower\graphtemp\hbox to 0pt{\hss {\footnotesize\em loc:}\hss}}%
        \graphtemp=.5ex\advance\graphtemp by 1.650in
        \rlap{\kern 2.131in\lower\graphtemp\hbox to 0pt{\hss {\sc the}\hss}}%
        \special{pa 1838 1300}%
        \special{pa 2131 1500}%
        \special{fp}%
        \special{sh 1.000}%
        \special{pa 2062 1423}%
        \special{pa 2131 1500}%
        \special{pa 2034 1464}%
        \special{pa 2062 1423}%
        \special{fp}%
        \graphtemp=.5ex\advance\graphtemp by 1.400in
        \rlap{\kern 1.985in\lower\graphtemp\hbox to 0pt{\hss {\footnotesize\em det:}}}%
        \hbox{\vrule depth1.800in width0pt height 0pt}%
        \kern 2.323in
        }%
      }%
    \centerline{\raise 1em\box\graph}
    \caption{\em ``The dog was running in the house''}
    \label{fig:wasrun}
  \end{center}
\end{figure}

\begin{figure}[tb]
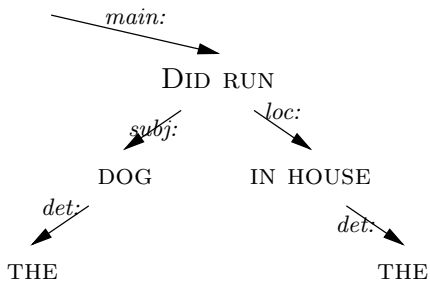

  \begin{center}
    \leavevmode
    \expandafter\ifx\csname graph\endcsname\relax \csname newbox\endcsname\graph\fi
    \expandafter\ifx\csname graphtemp\endcsname\relax \csname newdimen\endcsname\graphtemp\fi
    \setbox\graph=\vtop{\vskip 0pt\hbox{%
        \graphtemp=.5ex\advance\graphtemp by 1.650in
        \rlap{\kern 0.192in\lower\graphtemp\hbox to 0pt{\hss {\sc the}\hss}}%
        \graphtemp=.5ex\advance\graphtemp by 1.150in
        \rlap{\kern 0.677in\lower\graphtemp\hbox to 0pt{\hss {\sc dog}\hss}}%
        \special{pn 8}%
        \special{pa 485 1300}%
        \special{pa 192 1500}%
        \special{fp}%
        \special{sh 1.000}%
        \special{pa 289 1464}%
        \special{pa 192 1500}%
        \special{pa 261 1423}%
        \special{pa 289 1464}%
        \special{fp}%
        \graphtemp=\baselineskip\multiply\graphtemp by -1\divide\graphtemp by 2
        \advance\graphtemp by .5ex\advance\graphtemp by 1.400in
        \rlap{\kern 0.338in\lower\graphtemp\hbox to 0pt{\hss {\footnotesize\em det:}\hss}}%
        \graphtemp=.5ex\advance\graphtemp by 0.650in
        \rlap{\kern 1.162in\lower\graphtemp\hbox to 0pt{\hss {\sc Did run}\hss}}%
        \special{pa 292 300}%
        \special{pa 1162 500}%
        \special{fp}%
        \special{sh 1.000}%
        \special{pa 1070 453}%
        \special{pa 1162 500}%
        \special{pa 1058 502}%
        \special{pa 1070 453}%
        \special{fp}%
        \graphtemp=\baselineskip\multiply\graphtemp by -1\divide\graphtemp by 2
        \advance\graphtemp by .5ex\advance\graphtemp by 0.400in
        \rlap{\kern 0.727in\lower\graphtemp\hbox to 0pt{\hss {\footnotesize\em main:}\hss}}%
        \special{pa 969 800}%
        \special{pa 677 1000}%
        \special{fp}%
        \special{sh 1.000}%
        \special{pa 774 964}%
        \special{pa 677 1000}%
        \special{pa 745 923}%
        \special{pa 774 964}%
        \special{fp}%
        \graphtemp=.5ex\advance\graphtemp by 0.900in
        \rlap{\kern 0.823in\lower\graphtemp\hbox to 0pt{\hss {\footnotesize\em subj:}\hss}}%
        \graphtemp=.5ex\advance\graphtemp by 1.150in
        \rlap{\kern 1.646in\lower\graphtemp\hbox to 0pt{\hss {\sc in house}\hss}}%
        \special{pa 1354 800}%
        \special{pa 1646 1000}%
        \special{fp}%
        \special{sh 1.000}%
        \special{pa 1578 923}%
        \special{pa 1646 1000}%
        \special{pa 1550 964}%
        \special{pa 1578 923}%
        \special{fp}%
        \graphtemp=\baselineskip\multiply\graphtemp by -1\divide\graphtemp by 2
        \advance\graphtemp by .5ex\advance\graphtemp by 0.900in
        \rlap{\kern 1.500in\lower\graphtemp\hbox to 0pt{\hss {\footnotesize\em loc:}\hss}}%
        \graphtemp=.5ex\advance\graphtemp by 1.650in
        \rlap{\kern 2.131in\lower\graphtemp\hbox to 0pt{\hss {\sc the}\hss}}%
        \special{pa 1838 1300}%
        \special{pa 2131 1500}%
        \special{fp}%
        \special{sh 1.000}%
        \special{pa 2062 1423}%
        \special{pa 2131 1500}%
        \special{pa 2034 1464}%
        \special{pa 2062 1423}%
        \special{fp}%
        \graphtemp=.5ex\advance\graphtemp by 1.400in
        \rlap{\kern 1.985in\lower\graphtemp\hbox to 0pt{\hss {\footnotesize\em det:}}}%
        \hbox{\vrule depth1.800in width0pt height 0pt}%
        \kern 2.323in
        }%
      }%
    \centerline{\raise 1em\box\graph}
    \caption{\em ``Did the dog run in the house''}
    \label{fig:didrun}
  \end{center}
\end{figure}

To obtain a closer correspondence with the semantic structure, the
{\em nucleus format} corresponding to Tesni{\`e}re's stemmas is applied.
It is useful for many practical purposes.  Consider, for example,
collecting arguments for a given verb ``RUN''.  Having the analysis
such as those illustrated in Figure~\ref{fig:wasrun}, it is easy to
excerpt all sentences where the governing node is verbal having a
main element that has ``run'' as the base form, e.g.~{\em ran}, {\em ``was
  running''} (Figure~\ref{fig:wasrun}), {\em ``did run''}
(Figure~\ref{fig:didrun}).  The contraction form {\em ``won't run''}
obtains the same analysis (the same tree although the word nuclei can
contain extra information which makes the distinction) as a
contraction of the words {\em ``will not run''}.  As the example
shows, orthographic words were segmented whenever required by
the syntactic analysis.

This solution did not exist prior the FDG and generally is not
possible in a monostratal dependency description, which takes the
(orthographic) words as primitives.  The problem is that the
non-contiguous elements in a verb-chain are assigned into a single node 
while the subject in between belongs to its own node.

For historical reasons, the representation contains a
lexico-functional level closely similar to the syntactic analysis of
the earlier English Constraint Grammar (ENGCG) \cite{FKall95} parsing
system.  The current FDG formalism overcomes several
shortcomings\footnote{Listed in \newcite{AV94dpg}.} of the earlier
approaches: (1) the FDG does not rely on the detection of clause
boundaries, (2) parsing is no longer sequential, (3) ambiguity is
represented at the clause level rather than word level, (4) due to
explicit representation of dependency structure, there is no need to
refer to phrase-like units.  Because the FDG rule formalism is more
expressive, linguistic generalisation can be formalised in a more
transparent way, which makes the rules more readable.

\section{Descriptive solutions}
\subsection{Coordination}

We now tackle the problem of how coordination can be represented in
the framework of dependency model.  For example, \newcite{RH91} has
argued that coordination is a phenomenon that requires resorting to a
phrase-structure model.  

Coordination should not be seen as a directed functional relation, but
instead as a special connexion between two functionally equal
elements.  The coordination connexions are called junctions in
\newcite[Chs.~134-150]{LT59}.  Tesni{\`e}re considered junctions primarily
as a mechanism to pack multiple sentences economically into one.
Unfortunately, his solution, which represents all coordinative connexions
in stemmas, is not adequate, because due to cyclic arcs
the result is no longer a tree.

\begin{figure}[tb!]
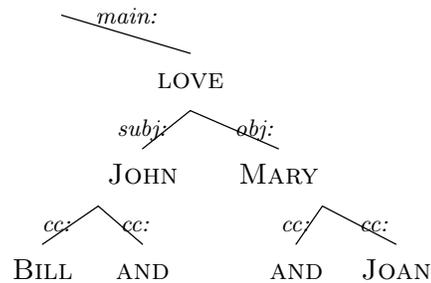

  \begin{center}
    \leavevmode
    \expandafter\ifx\csname graph\endcsname\relax \csname newbox\endcsname\graph\fi
    \expandafter\ifx\csname graphtemp\endcsname\relax \csname newdimen\endcsname\graphtemp\fi
    \setbox\graph=\vtop{\vskip 0pt\hbox{%
        \graphtemp=.5ex\advance\graphtemp by 0.150in
        \rlap{\kern 0.231in\lower\graphtemp\hbox to 0pt{\hss {\sc\small }\hss}}%
        \graphtemp=.5ex\advance\graphtemp by 1.650in
        \rlap{\kern 0.231in\lower\graphtemp\hbox to 0pt{\hss {\sc Bill}\hss}}%
        \graphtemp=.5ex\advance\graphtemp by 1.650in
        \rlap{\kern 0.754in\lower\graphtemp\hbox to 0pt{\hss {\sc and}\hss}}%
        \graphtemp=.5ex\advance\graphtemp by 1.150in
        \rlap{\kern 0.754in\lower\graphtemp\hbox to 0pt{\hss {\sc John}\hss}}%
        \special{pn 8}%
        \special{pa 523 1300}%
        \special{pa 231 1500}%
        \special{fp}%
        \graphtemp=.5ex\advance\graphtemp by 1.400in
        \rlap{\kern 0.377in\lower\graphtemp\hbox to 0pt{\hss {\footnotesize\em cc:}}}%
        \special{pa 523 1300}%
        \special{pa 754 1500}%
        \special{fp}%
        \graphtemp=.5ex\advance\graphtemp by 1.400in
        \rlap{\kern 0.638in\lower\graphtemp\hbox to 0pt{{\footnotesize\em cc:}\hss}}%
        \graphtemp=.5ex\advance\graphtemp by 0.650in
        \rlap{\kern 1.004in\lower\graphtemp\hbox to 0pt{\hss {\sc love}\hss}}%
        \special{pa 331 300}%
        \special{pa 1004 500}%
        \special{fp}%
        \graphtemp=\baselineskip\multiply\graphtemp by -1\divide\graphtemp by 2
        \advance\graphtemp by .5ex\advance\graphtemp by 0.400in
        \rlap{\kern 0.667in\lower\graphtemp\hbox to 0pt{\hss {\footnotesize\em main:}\hss}}%
        \special{pa 1004 800}%
        \special{pa 754 1000}%
        \special{fp}%
        \graphtemp=.5ex\advance\graphtemp by 0.900in
        \rlap{\kern 0.879in\lower\graphtemp\hbox to 0pt{\hss {\footnotesize\em subj:}}}%
        \graphtemp=.5ex\advance\graphtemp by 1.150in
        \rlap{\kern 1.465in\lower\graphtemp\hbox to 0pt{\hss {\sc Mary}\hss}}%
        \special{pa 1004 800}%
        \special{pa 1465 1000}%
        \special{fp}%
        \graphtemp=.5ex\advance\graphtemp by 0.900in
        \rlap{\kern 1.235in\lower\graphtemp\hbox to 0pt{{\footnotesize\em obj:}\hss}}%
        \graphtemp=.5ex\advance\graphtemp by 1.650in
        \rlap{\kern 1.558in\lower\graphtemp\hbox to 0pt{\hss {\sc and}\hss}}%
        \special{pa 1696 1300}%
        \special{pa 1558 1500}%
        \special{fp}%
        \graphtemp=.5ex\advance\graphtemp by 1.400in
        \rlap{\kern 1.627in\lower\graphtemp\hbox to 0pt{\hss {\footnotesize\em cc:}}}%
        \graphtemp=.5ex\advance\graphtemp by 1.650in
        \rlap{\kern 2.081in\lower\graphtemp\hbox to 0pt{\hss {\sc Joan}\hss}}%
        \special{pa 1696 1300}%
        \special{pa 2081 1500}%
        \special{fp}%
        \graphtemp=.5ex\advance\graphtemp by 1.400in
        \rlap{\kern 1.888in\lower\graphtemp\hbox to 0pt{{\footnotesize\em cc:}\hss}}%
        \hbox{\vrule depth1.800in width0pt height 0pt}%
        \kern 2.312in
        }%
      }%
    \centerline{\raise 1em\box\graph}
    \caption{Coordinated elements}
    \label{fig:coord}
  \end{center}
\end{figure}

Our solution is to pay due respect to the formal properties of the dependency
model, which requires that each element should have one and only one
head.\footnote{The treatment of coordination and gapping in \newcite{SK97}
  resembles ours in simple cases.  However, this model maintains projectivity,
  and consequently, both multiple heads and extended nuclei, which are
  essentially phrase-level units, are used in complex cases, making the model
  broadly similar to \newcite{RH91}.} This means that coordinated elements are
chained (Figure~\ref{fig:coord}) using a specific arc for coordination
(labeled as {\em cc\/}).  The coordinators are mostly redundant markers
\cite[Ch.~39:5]{LT59}\footnote{The redundancy is shown in the existence of
  asyndetic coordination. As syntactic markers, coordinators are not
  completely void of semantic content, which is demonstrated by the existence
  of contrasting set of coordinators; 'and', 'or', 'but' etc.}, especially,
they do not have any (governing) role in the syntactic structure as they do in
many word-based forms of dependency theory (e.g.  \newcite{JK75} and
\newcite{IM87}).

Unlike the other arcs in the tree, the arc marking coordination does
not imply a dependency relation but rather a functional equivalence.
If we assume that the coordinated elements have exactly the same
syntactic functions, the information available is similar to that provided
in Tesni{\`e}re's representation.  If needed, we can simply print all the
possible combinations of the coordinated elements: {\em ``Bill loves
  Mary''}, {\em ``John loves Mary''}, etc.

\subsection{Gapping}

It is claimed that gapping is even a more serious problem for
dependency theories, a phenomenon which requires the presence of
non-terminal nodes.  The treatment of gapping, where the main verb of
a clause is missing, follows from the treatment of simple coordination.

\begin{figure*}
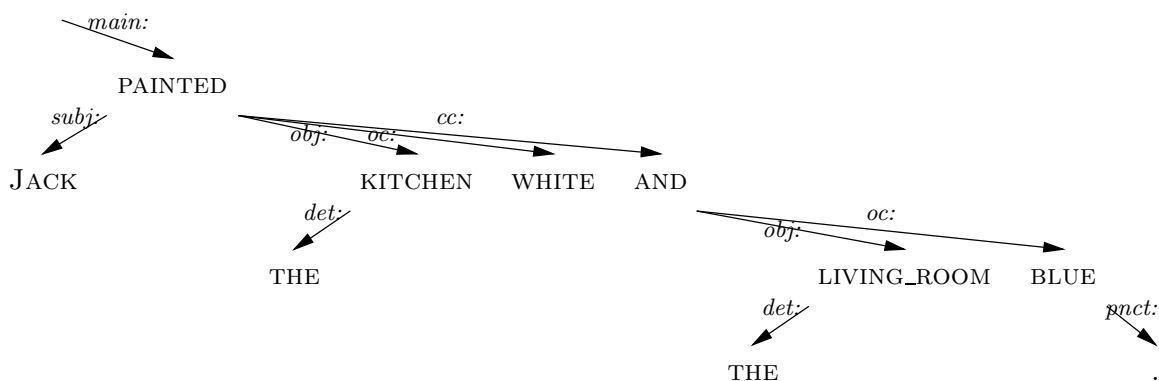

  \begin{center}
    \leavevmode
    \expandafter\ifx\csname graph\endcsname\relax \csname newbox\endcsname\graph\fi
    \expandafter\ifx\csname graphtemp\endcsname\relax \csname newdimen\endcsname\graphtemp\fi
    \setbox\graph=\vtop{\vskip 0pt\hbox{%
        \graphtemp=.5ex\advance\graphtemp by 0.150in
        \rlap{\kern 0.231in\lower\graphtemp\hbox to 0pt{\hss {\sc\small }\hss}}%
        \graphtemp=.5ex\advance\graphtemp by 1.150in
        \rlap{\kern 0.231in\lower\graphtemp\hbox to 0pt{\hss {\sc Jack}\hss}}%
        \graphtemp=.5ex\advance\graphtemp by 0.650in
        \rlap{\kern 0.908in\lower\graphtemp\hbox to 0pt{\hss {\sc painted}\hss}}%
        \special{pn 8}%
        \special{pa 331 300}%
        \special{pa 908 500}%
        \special{fp}%
        \special{sh 1.000}%
        \special{pa 821 444}%
        \special{pa 908 500}%
        \special{pa 805 491}%
        \special{pa 821 444}%
        \special{fp}%
        \graphtemp=\baselineskip\multiply\graphtemp by -1\divide\graphtemp by 2
        \advance\graphtemp by .5ex\advance\graphtemp by 0.400in
        \rlap{\kern 0.619in\lower\graphtemp\hbox to 0pt{\hss {\footnotesize\em main:}\hss}}%
        \special{pa 562 800}%
        \special{pa 231 1000}%
        \special{fp}%
        \special{sh 1.000}%
        \special{pa 329 970}%
        \special{pa 231 1000}%
        \special{pa 303 927}%
        \special{pa 329 970}%
        \special{fp}%
        \graphtemp=\baselineskip\multiply\graphtemp by -1\divide\graphtemp by 2
        \advance\graphtemp by .5ex\advance\graphtemp by 0.900in
        \rlap{\kern 0.396in\lower\graphtemp\hbox to 0pt{\hss {\footnotesize\em subj:}\hss}}%
        \graphtemp=.5ex\advance\graphtemp by 1.650in
        \rlap{\kern 1.546in\lower\graphtemp\hbox to 0pt{\hss {\sc the}\hss}}%
        \graphtemp=.5ex\advance\graphtemp by 1.150in
        \rlap{\kern 2.185in\lower\graphtemp\hbox to 0pt{\hss {\sc kitchen}\hss}}%
        \special{pa 1254 800}%
        \special{pa 2185 1000}%
        \special{fp}%
        \special{sh 1.000}%
        \special{pa 2092 955}%
        \special{pa 2185 1000}%
        \special{pa 2082 1003}%
        \special{pa 2092 955}%
        \special{fp}%
        \graphtemp=.5ex\advance\graphtemp by 0.900in
        \rlap{\kern 1.719in\lower\graphtemp\hbox to 0pt{\hss {\footnotesize\em obj:}}}%
        \special{pa 1838 1300}%
        \special{pa 1546 1500}%
        \special{fp}%
        \special{sh 1.000}%
        \special{pa 1643 1464}%
        \special{pa 1546 1500}%
        \special{pa 1615 1423}%
        \special{pa 1643 1464}%
        \special{fp}%
        \graphtemp=\baselineskip\multiply\graphtemp by -1\divide\graphtemp by 2
        \advance\graphtemp by .5ex\advance\graphtemp by 1.400in
        \rlap{\kern 1.692in\lower\graphtemp\hbox to 0pt{\hss {\footnotesize\em det:}\hss}}%
        \graphtemp=.5ex\advance\graphtemp by 1.150in
        \rlap{\kern 2.900in\lower\graphtemp\hbox to 0pt{\hss {\sc white}\hss}}%
        \special{pa 1254 800}%
        \special{pa 2900 1000}%
        \special{fp}%
        \special{sh 1.000}%
        \special{pa 2804 963}%
        \special{pa 2900 1000}%
        \special{pa 2798 1013}%
        \special{pa 2804 963}%
        \special{fp}%
        \graphtemp=.5ex\advance\graphtemp by 0.900in
        \rlap{\kern 2.077in\lower\graphtemp\hbox to 0pt{\hss {\footnotesize\em oc:}}}%
        \graphtemp=.5ex\advance\graphtemp by 1.150in
        \rlap{\kern 3.462in\lower\graphtemp\hbox to 0pt{\hss {\sc and}\hss}}%
        \special{pa 1254 800}%
        \special{pa 3462 1000}%
        \special{fp}%
        \special{sh 1.000}%
        \special{pa 3364 966}%
        \special{pa 3462 1000}%
        \special{pa 3360 1016}%
        \special{pa 3364 966}%
        \special{fp}%
        \graphtemp=\baselineskip\multiply\graphtemp by -1\divide\graphtemp by 2
        \advance\graphtemp by .5ex\advance\graphtemp by 0.900in
        \rlap{\kern 2.358in\lower\graphtemp\hbox to 0pt{\hss {\footnotesize\em cc:}\hss}}%
        \graphtemp=.5ex\advance\graphtemp by 2.150in
        \rlap{\kern 3.946in\lower\graphtemp\hbox to 0pt{\hss {\sc the}\hss}}%
        \graphtemp=.5ex\advance\graphtemp by 1.650in
        \rlap{\kern 4.738in\lower\graphtemp\hbox to 0pt{\hss {\sc living\_room}\hss}}%
        \special{pa 3654 1300}%
        \special{pa 4738 1500}%
        \special{fp}%
        \special{sh 1.000}%
        \special{pa 4645 1457}%
        \special{pa 4738 1500}%
        \special{pa 4636 1506}%
        \special{pa 4645 1457}%
        \special{fp}%
        \graphtemp=.5ex\advance\graphtemp by 1.400in
        \rlap{\kern 4.196in\lower\graphtemp\hbox to 0pt{\hss {\footnotesize\em obj:}}}%
        \special{pa 4238 1800}%
        \special{pa 3946 2000}%
        \special{fp}%
        \special{sh 1.000}%
        \special{pa 4043 1964}%
        \special{pa 3946 2000}%
        \special{pa 4015 1923}%
        \special{pa 4043 1964}%
        \special{fp}%
        \graphtemp=\baselineskip\multiply\graphtemp by -1\divide\graphtemp by 2
        \advance\graphtemp by .5ex\advance\graphtemp by 1.900in
        \rlap{\kern 4.092in\lower\graphtemp\hbox to 0pt{\hss {\footnotesize\em det:}\hss}}%
        \graphtemp=.5ex\advance\graphtemp by 1.650in
        \rlap{\kern 5.569in\lower\graphtemp\hbox to 0pt{\hss {\sc blue}\hss}}%
        \special{pa 3654 1300}%
        \special{pa 5569 1500}%
        \special{fp}%
        \special{sh 1.000}%
        \special{pa 5472 1465}%
        \special{pa 5569 1500}%
        \special{pa 5467 1514}%
        \special{pa 5472 1465}%
        \special{fp}%
        \graphtemp=\baselineskip\multiply\graphtemp by -1\divide\graphtemp by 2
        \advance\graphtemp by .5ex\advance\graphtemp by 1.400in
        \rlap{\kern 4.612in\lower\graphtemp\hbox to 0pt{\hss {\footnotesize\em oc:}\hss}}%
        \graphtemp=.5ex\advance\graphtemp by 2.150in
        \rlap{\kern 6.054in\lower\graphtemp\hbox to 0pt{\hss {\sc .}\hss}}%
        \special{pa 5800 1800}%
        \special{pa 6054 2000}%
        \special{fp}%
        \special{sh 1.000}%
        \special{pa 5991 1918}%
        \special{pa 6054 2000}%
        \special{pa 5960 1958}%
        \special{pa 5991 1918}%
        \special{fp}%
        \graphtemp=\baselineskip\multiply\graphtemp by -1\divide\graphtemp by 2
        \advance\graphtemp by .5ex\advance\graphtemp by 1.900in
        \rlap{\kern 5.927in\lower\graphtemp\hbox to 0pt{\hss {\footnotesize\em pnct:}\hss}}%
        \hbox{\vrule depth2.300in width0pt height 0pt}%
        \kern 6.208in
        }%
      }%
    \centerline{\raise 1em\box\graph}
    \caption{{\em Jack painted the kitchen white and the living room blue.}}
    \label{fig:973}
  \end{center}
\end{figure*}

In simple coordination, the coordinator has an auxiliary role without any
specific function in the syntactic tree.  In gapping, only the coordinator
is present while the verb is missing.  One can think that as the
coordinator represents all missing elements in the clause, it inherits all
properties of the missing (verbal) elements (Figure~\ref{fig:973}).
This solution is also computationally effective because we do not need to
postulate empty nodes in the actual parsing system.

 From a descriptive point of view there is no problem if we think that
the coordinator obtains syntactic properties from the nucleus that it
is connected to.  Thus, in a sentence with verbal ellipsis, e.g.~in
the sentence {\em ``Jack painted the kitchen white and the living room
  blue''}, the coordinator obtains the subcategorisation properties of
a verb.  A corresponding graph is seen in Figure~\ref{fig:973}.

Due to 'flatness' of dependency model, there is no problem to describe gapping
where a subject rather than complements are involved, as the
Figure~\ref{fig:CG-style} shows.  Note that gapping provides clear evidence
that the syntactic element is a {\em nucleus} rather than a word.  For
example, in the sentence {\em ``Jack has been lazy and Jill angry''}, the
elliptic element is the verbal nucleus {\em has been}.

\begin{figure}[tb]
  \begin{center}
    \leavevmode
    \begin{tabbing}\sf\small
  $<$m \= \kill
  $<$John$>$ \\
    \> "John" N SG @SUBJ~subj:$>$2 \\
  $<$gave$>$ \\
    \> "give" V PAST @+FV~\#2~main:$>$0 \\
  $<$the$>$ \\
    \> "the" DET ART SG/PL @DN$>$~det:$>$4 \\
  $<$lecture$>$ \\
    \> "lecture" N SG @OBJ \#4~obj:$>$2 \\
  $<$on$>$ \\
    \> "on" PREP @ADVL \#5~tmp:$>$2 \\
  $<$Tuesday$>$ \\
    \> "Tuesday" N SG @$<$P~pcomp:$>$5 \\
  $<$and$>$ \\
    \> "and" CC @CC \#7~cc:$>$2 \\
  $<$Bill$>$ \\
    \> "Bill" N SG @SUBJ subj:$>$7 \\
  $<$on$>$ \\
    \> "on" PREP @ADVL \#9~tmp:$>$7 \\
  $<$Wednesday$>$ \\
    \> "Wednesday" N SG @$<$P~pcomp:$>$9 \\
  $<$.$>$ 
    \end{tabbing}
    \caption{Text-based representation}
    \label{fig:CG-style}
  \end{center}
\end{figure}

\section{Conclusion}

This paper argues for a descriptively adequate syntactic theory that
is based on dependency rather than constituency.  Tesni{\`e}re's theory
seems to provide a useful descriptive framework for syntactic phenomena
occurring in various natural languages.  We apply the theory and
develop the representation to meet the requirements of computerised
parsing description.  Simultaneously, we explicate the formal
properties of Tesni{\`e}re's theory that are used in constructing a
practical parsing system.

A solution to the main obstacle to the utilisation of the theory, the
linearisation of the syntactic structure, is presented.  As a case
study, we reformulate the theory for the description of coordination
and gapping, which are difficult problems for any comprehensive
syntactic theory.

\vspace{2pt}
\begin{acknowledgments}
  We thank Fred Karlsson, Atro Voutilainen and three Coling-ACL '98 workshop
 referees for useful comments on earlier draft of this paper.
\end{acknowledgments}

\end{document}